\documentclass[lettersize,journal]{IEEEtran}
\usepackage{amsmath,amsfonts}
\usepackage{algorithmic}
\usepackage{algorithm}
\usepackage{array}
\usepackage[caption=false,font=normalsize,labelfont=sf,textfont=sf]{subfig}
\usepackage{textcomp}
\usepackage{stfloats}
\usepackage{url}
\usepackage{verbatim}
\usepackage{graphicx}
\usepackage[utf8]{inputenc}
\usepackage{cite}
\title{Hybrid DQN-TD3 Reinforcement Learning for Autonomous Navigation in Dynamic Environments}
\author{Xiaoyi He, Danggui Chen, Zhenshuo Zhang, Zimeng Bai}
\date{\today}
\begin{document}
\maketitle 
\begin    {abstract}
This paper proposes a hierarchical path planning and control framework that integrates the strategic decision-making capability of Deep Q-Network (DQN) for discrete subgoal selection with the precise continuous control execution of Twin Delayed Deep Deterministic Policy Gradient (TD3). By leveraging DQN's strength in high-level discrete decision-making and TD3's sample efficiency and stability in continuous motion domains, the proposed approach achieves robust policy generalization and enhanced adaptive performance in dynamic and uncertain environments, effectively overcoming limitations of conventional single-algorithm solutions in complex robotic applications.
\end{abstract}


\section{INTRODUCTION}

\IEEEPARstart{T}{raditional} path planning algorithms, such as A* and Dijkstra, are predicated on prior environmental modeling (i.e. pre-constructed maps or graph structures). While they demonstrate efficacy in static environments, they exhibit inherent limitations in dynamic and unstructured scenarios. In such contexts, frequent global path replanning is necessitated, resulting in exponential degradation of computational efficiency\cite{zhang2014multiple}, coupled with non-negligible latency and performance overhead. 

On the other hand, single reinforcement learning methods have their own limitations. Deep Q-Network (DQN) excels at making discrete decisions such as direction or path selection, but lacks the ability to handle fine-grained continuous control. Twin Delayed Deep Deterministic Policy Gradient (TD3) performs well in continuous action spaces, offering stable and efficient control, but is less effective in handling high-level, discrete navigation strategies.\cite{hausknecht2017parametrized}.

This research aims to develop a hybrid reinforcement learning architecture that combines DQN for high-level decision-making and TD3 for continuous control. The framework is designed to enhance navigation accuracy and obstacle avoidance in dynamic environments, ultimately achieving an adaptive and efficient autonomous navigation system.

This work draws inspiration from hierarchical reinforcement learning (HRL) literature, where high-level and low-level policies often share a unified reward structure for consistent optimization\cite{fan2019hybrid}. Our proposed DQN-TD3 hybrid control framework employs a unified reward mechanism to simultaneously drive high-level strategic decision-making and low-level continuous motion control. Additionally, prior research in deep RL-based navigation, such as Tai et al.~\cite{tai2017virtual} and Pfeiffer et al.~\cite{pfeiffer2017perception}, has demonstrated the effectiveness of multi-objective reward designs encompassing goal achievement, safety, and control smoothness, which further informs our reward.

\subsection{Practical Implication}
In real-world applications, mobile robots are widely used in warehouse logistics, indoor service, security patrol, and search-and-rescue tasks. These environments are often highly dynamic, featuring moving obstacles (e.g., humans or other robots), incomplete or constantly changing map information, and conditions where GPS signals are weak or entirely unavailable.

Traditional path planning methods such as A* and Dijkstra rely on static map
s and deterministic graph structures, making them poorly suited for fast adaptation in changing environments. However, real-world scenarios demand that robots possess the ability to perceive, decide, and act autonomously in real time, even in unknown or partially observable environments.

Therefore, developing an adaptive navigation method based on reinforcement learning not only enhances the robot's robustness and environmental adaptability but also addresses key limitations of classical planners, such as map dependency and limited reactivity. This is crucial for enabling reliable and intelligent robotic systems in complex, dynamic, and uncertain real-world settings.
\subsection{Current solutions and gaps}
Traditional planners provide deterministic solutions but lack adaptability. To improve the performance of automatic navigation in dynamic and complex environments, prior works have applied deep reinforcement learning (DRL), particularly single algorithms such as DQN or TD3\cite{hausknecht2017parametrized}. Existing RL-based methods improve adaptability and efficiency but have inherent limitations: DQN cannot directly output continuous control commands, while TD3 requires careful tuning and strategy development, especially in tasks involving dynamic environments.\cite{fujimoto2018addressing}.

\section{RELATED WORK}
With growing awareness, mobile robot automatic navigation has emerged as a significant research area of robotics, and DRL methods for continuous control have gained substantial attention in recent years. This section provides an overview of relevant literature and highlights important contributions as well as gaps in the field.

\subsection{Prior Work}
For a long time, path planning has attracted extensive attention in the field of artificial intelligence. Many classical algorithms such as Dijkstra algorithm, A* algorithm, Simulated Annealing (SA) and Ant Colony Optimization (ACO) have appealed. With elevated criteria for a complex and dynamic environment, RL is required in path planning. Dynamic programming, Q-learning, SARSA and other reinforcement algorithms have been proposed, but these tabular reinforcement learning methods have obvious limitations in the size of state space and action space\cite{Priorwork1}.

Researchers then applied DRL, particularly DQN, which effectively utilizes a neural network \cite{Priorwork2}. Enhancements based on DQN, such as the dueling network structure and Double DQN (DDQN) \cite{wang2016dueling}, addressed critical issues like Q-value overestimation and improved the stability of DQN. PPO also has great potential in path planning, advancing policy gradient methods \cite{Priorwork9,Priorwork10}. Meanwhile, TD3 demonstrates superior performance in continuous control tasks.\cite{fujimoto2018addressing}

\subsection{Missing Part in Mobile Robot Navigation}
Single-RL methods (DQN and TD3) have achieved success in constrained or structured environments.\cite{mnih2015human}\cite{fujimoto2018addressing} Despite recent progress, the integration of DQN and PPO is not entirely uncharted\cite{hausknecht2017parametrized}, but exploration in the integration of two RL algorithms (e.g. DQN and TD3) holds novel potential in automatic navigation. This combined approach receives little attention in research, which is expected to achieve superior performance and mitigate constraints imposed by a single algorithm.

\subsection{Novelty and Advantages of the Proposed Framework}
This paper integrates DQN’s capacity for discrete topological decision-making with TD3’s proficiency in fine-grained continuous control, aiming to achieve robust policy generalization and enhanced adaptive performance in dynamic and partially observable environments.

\section{Methodology Sketch}

\subsection{Pipeline Diagram Description}
    \begin{figure*}[t]
    \centering
    \includegraphics[width=0.8\linewidth]{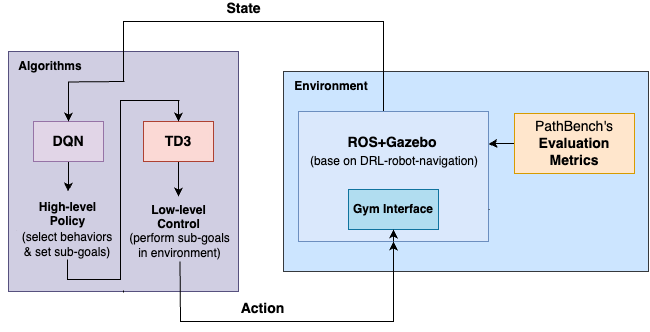}
    \caption{This is the pipeline diagram description.}
    \label{Pipeline Diagram Description}
    \end{figure*}
As illustrated in Figure\ref{Pipeline Diagram Description}, this research is dedicated to developing a hybrid policy model combining DQN and TD3. We implement the algorithm in a PyBullet\cite{coumans2021} simulated robot and evaluate its performance on metrics (like path optimality, collision rate, and re-planning efficiency) with comparative baselines.

\subsection{Algorithm Modules}
Inspired by hierarchical reinforcement learning (HRL), we bridge the gap between high-level strategic decision-making and low-level continuous control by designing a unified reward mechanism that is compatible with both value-based methods (DQN) and policy-gradient algorithms (TD3). Our design draws from prior works that emphasize multiple objectives in autonomous navigation, including goal achievement, safety, and motion efficiency \cite{tai2017virtual, pfeiffer2017perception, chen2017decentralized}.

We define the high-level agent reward function as:

\begin{equation}
\begin{aligned}
R_{high} &= w_1 \cdot R_{dir} + w_2 \cdot R_{dist} + w_3 \cdot R_{avoid} \\
        &\quad+ w_4 \cdot R_{smooth} - P_{collision} - P_{time}
\end{aligned}
\label{eq:reward}
\end{equation}

Where:

\begin{itemize}
  \item \textbf{Direction reward}:
  \begin{equation}
  R_{dir} = 1 - \frac{|\theta_{diff}|}{180}
  \end{equation}
  encourages the agent to move towards the target direction. Here, $\theta_{diff}$ is the angle difference between the actual direction and the target direction.

  \item \textbf{Distance reward}:
  \begin{equation}
  R_{dist} = 1 - \min\left(\frac{|d_{actual} - d_{target}|}{d_{target}}, 1\right)
  \end{equation}
  encourages the agent to approach the target location. $d_{actual}$ is the actual distance moved by the agent, and $d_{target}$  is the target distance.

  \item \textbf{Obstacle Avoidance reward}:
  \begin{equation}
  R_{avoid} = \begin{cases}
    +r_{\text{avoidance}} & \text{if there is no obstacle ahead} \\
    0 & \text{otherwise}
    \end{cases}
  \end{equation}
  encourages the agent to avoid obstacles.

  \item \textbf{Path Smoothness reward}:
  \begin{equation}
  R_{smooth} = 0.1 \cdot \left(1 - \min\left(\frac{|\Delta\theta|}{90}, 1\right)\right)
  \end{equation}
  encourages the agent to minimize directional changes between actions. $\Delta\theta$ is the change in direction between consecutive actions.

  \item \textbf{Collision penalty}:
  \begin{equation}
  P_{\text{collision}} =
  \begin{cases}
  +p_{\text{collision}}, & \text{if a collision occurs} \\
  0, & \text{otherwise}
  \end{cases}
  \end{equation}
  penalizes the agent for collisions.

  \item \textbf{Time Penalty}:
  \begin{equation}
  P_{time} = p_{\text{time}}
  \end{equation}
  encourages the agent to reduce task completion time.
\end{itemize}

The weights for each component in the code are:
\begin{equation}
  w_1 = 0.4, w_2 = 0.4, w_3 = 0.1, w_4 = 0.1
\end{equation}
\begin{equation}
  r_{\text{avoidance}}= 0.2, p_{\text{collision}}= 1.0, 
  p_{time} = 0.01
\end{equation}

The reward function for the low-level agent can be expressed as:

\begin{equation}
R_{low} = R_{env} + w_7 \cdot (R_{dir} + R_{dist}) - w_8 \cdot P_{collision}
\end{equation}

Where:
\begin{itemize}

    \item \textbf{Environment reward}:
    \begin{equation}
      R_{env} =
    \begin{cases}
      100.0, & \text{if reached target }  \\
      -100.0, & \text{if collision occurs }  \\
      \frac{a_{lin}}{2} - \frac{|a_{ang}|}{2} - \frac{r(d_{min})}{2}, & \text{otherwise}
    \end{cases}
    \end{equation}
    encourages the agent to approach the target location. Where
    
    $a_{lin}$ : The robot’s current linear velocity 

    $a_{ang}$ : The robot’s current angular velocity 
    
    $d_{min}$ : The distance from the robot to the nearest obstacle
    
    $r(x) =
        \begin{cases}
          1 - x, & x < 1 \\
          0, & x \geq 1
        \end{cases}$

    \item \textbf{Subgoal completion reward}:
    \begin{equation}
      R_{dir} + R_{dist}
    \end{equation}
    encourages the low-level agent to complete subgoals set by the high-level agent.

    \item \textbf{Collision penalty}:
    \begin{equation}
      P_{collision}
    \end{equation}
    additionally penalizes the low-level agent for collisions.
      
\end{itemize}

This hierarchical framework facilitates the integration of long-term strategic planning (e.g., selecting optimal waypoints) and short-term reactive control (e.g., dynamic obstacle avoidance). The high-level DQN planner leverages reward signals for state-action value estimation, while the low-level TD3 controller utilizes reward gradients for precise policy optimization in continuous motion tasks.

\label{eq:q_learning_update}
\label{eq:policy_gradient}
\subsection{Environments}

In this project, we primarily utilize the Gazebo environment with ROS1 while integrating selected analyzer functionalities from PathBench\cite{toma2021pathbench} for evaluation purposes. Additional resources include the OpenAI Gymnasium (Gym)\cite{towers2024gymnasium} interface.

\vspace{1em} 
Gazebo serves as a powerful 3D robotics simulation environment, offering high-fidelity physics and sensor modeling. Integrated with ROS1 Noetic, it enables seamless development, testing, and deployment of robotic algorithms within realistic and complex environments. Additionally, Rviz is utilized for real-time visualization of robot states and sensor data, supporting advanced experimentation for reinforcement learning and motion planning.

We build upon DRL-robot-navigation\cite{9645287}, a repository for mobile robot navigation in the Gazebo simulator. The framework provides pre-configured robot models, Gazebo environment setups, and essential utilities including map construction and graph-based path planning algorithms, streamlining the development of our environment. Training is performed in the ROS Gazebo simulator with PyTorch, ROS Noetic on Ubuntu. Our primary implementation task involves creating a custom Gym interface for Gazebo environment to facilitate DRL algorithm training within this physical simulation environment.

\vspace{1em} 
OpenAI Gym establishes the API standard for reinforcement learning, offering a versatile interface that can be easily implemented across various Python environments while being capable of representing general RL problems. Since Stable-baselines3's DRL algorithms require Gym environments for training, our implementation ensures full compatibility with this standard, defining appropriate observation spaces, action spaces, and reward structures for the motion planning task.

\subsection{Benchmarks}
To rigorously evaluate the performance of our hierarchical reinforcement learning (HRL) framework, we employ a comprehensive set of benchmarking metrics tailored for motion planning in autonomous robotic systems. Key evaluation criteria include path optimality, computational efficiency, success rate, and smoothness of the generated trajectories. In our HRL architecture, the high-level planner utilizes the DQN algorithm for strategic decision-making, while the low-level controller employs the TD3 algorithm for precise motion execution. In our implementation, we adopt a TD3 implementation based on Stable-baselines3 as the baseline method, with all experimental evaluations conducted within the ROS-GAZEBO simulation platform.

\section{EXPERIMENT}
\subsection{Experiment Setup}
    \textbf{Simulation Environment and Hardware}
    
We test the capabilities of hybrid DRL (DQN and TD3) with the ROS-GAZEBO simulation environment, leveraging pytorch and tensorboard. All training and simulation experiments were conducted in the Gazebo simulator, with ROS1 Noetic and RViz for visualization. We use Docker as a containerization platform. Note that this project does not mainly rely on GPU acceleration due to high environment I/O overhead, which limits GPU utilization efficiency. 

In our experiment, training is conducted for approximately 10,000 episodes (5 million timesteps).
Each training episode was terminated under one of three conditions: the robot successfully reached the goal, a collision occurred, or a maximum of 500 time steps was consumed.

The maximum linear and angular velocities of the robot were set to [0, 1] m/s and [-1, 1] rad/s, respectively. The neural network parameters are updated every 100 timesteps for both high-level and low-level agents to ensure training stability.

    \textbf{Evaluation Metrics}
    
1) Success rate (goal reached).

	We will ensure that a valid path exists in the current map. When a path exists, this metric measures the probability of successfully reaching the goal point within the maximum time limit.
   
2) Collision rate.

	With realistic velocity conditions in the simulation environment, DRL models may still collide with obstacles. We aim to minimize the collision rate through adjustments to the reward function.

3) Time cost to goal point.

	For cases where the goal is successfully reached, we will measure the time cost across different algorithms, including time spent on path planning, time to reach the goal, and other relevant factors.
    
4) Path efficiency (distance/optimal).

	We will use the Euclidean distance as a reference and compare it with the efficiency of paths planned by the models.

5) Trajectory Smoothness.

        This metric evaluates the smoothness of the trajectory, reflecting the overall path quality.

\section{Results \& Discussion}
\label{sec:results_discussion}

\subsection{TD3 Training (Quantitative \& Qualitative)}
    \begin{figure}
    \centering
    \includegraphics[width=1\linewidth]{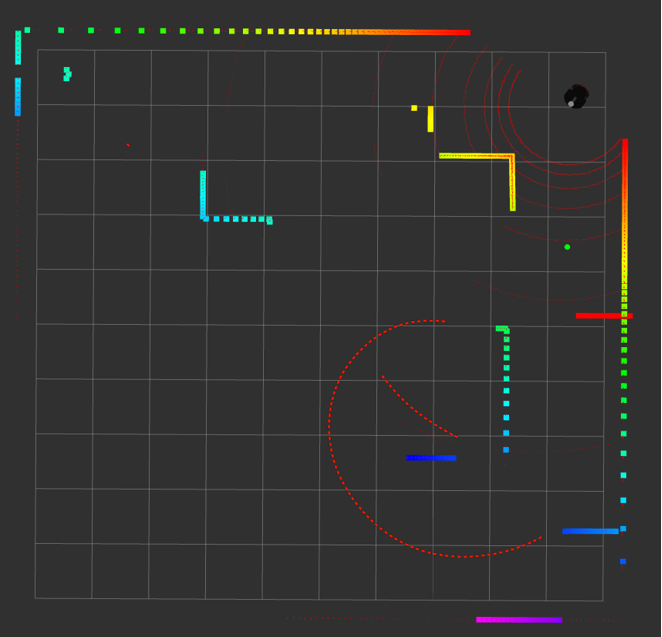}
    \caption{The RViz interface of the TD3 training process.}
    \label{fig:placeholder}
    \end{figure}
\textbf{Reward Curve:}
\begin{itemize}
\begin{figure}
    \centering
    \includegraphics[width=1\linewidth]{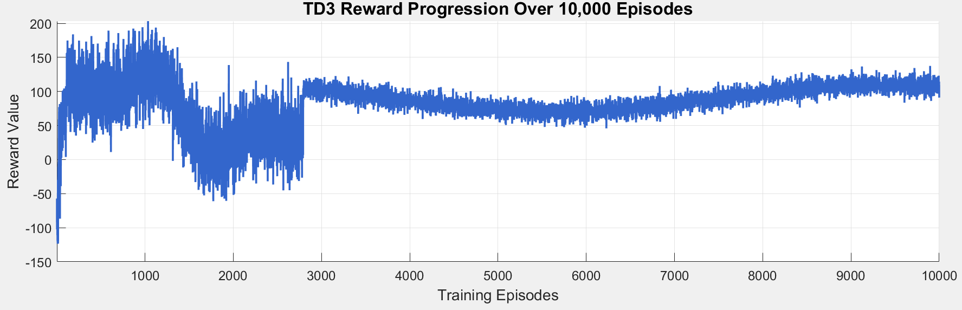}
    \caption{TD3 reward curve analysis.}
    \label{fig:placeholder}
\end{figure}
    \item Initial stage (Episodes 1--100): Reward rapidly rises from negative to positive values, indicating that the agent begins learning effective strategies.
    \item Mid stage (Episodes 100--2700): Fluctuates within 40--120, showing a balance between exploration and exploitation.
    \item Late stage (Episodes 2700--10000): Fluctuations decrease and stabilize around 80--110, strategy becomes mature.
    \begin{figure}
    \centering
    \includegraphics[width=1\linewidth]{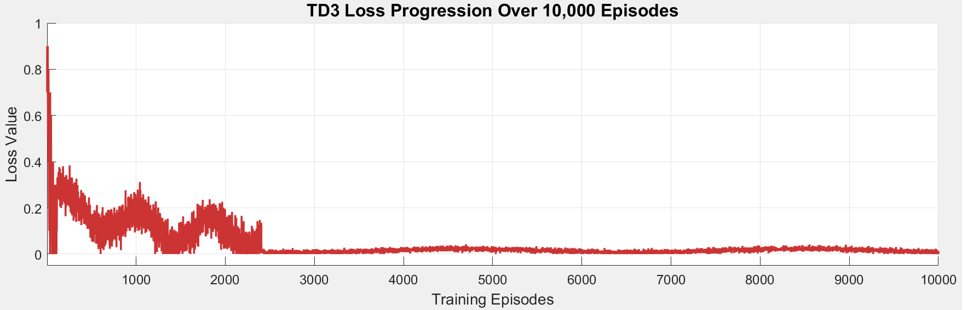}
    \caption{TD3 loss curve analysis.}
    \label{fig:placeholder}
    \end{figure}
\end{itemize}
\textbf{Loss Curve:}
\begin{itemize}

    \item Initial rapid decline (Episodes 1--110): Loss drops from $\sim$0.7--0.9 to 0.1--0.3, indicating rapid optimization of model parameters.
    \item Mid-term minor fluctuations (Episodes 110--2400): Loss fluctuates within 0.05--0.35, indicating continuous adjustment.
    \item Late-stage stable convergence (Episodes 2400--10000): Loss approaches zero (0.01--0.03), training stabilizes.
\end{itemize}

\textbf{Overall:} TD3 converges effectively but relatively slowly; curves smooth out after 3000 episodes.

\subsection{DQN+TD3 Experiment (Qualitative Observation)}
\begin{figure}
    \centering
    \includegraphics[width=1\linewidth]{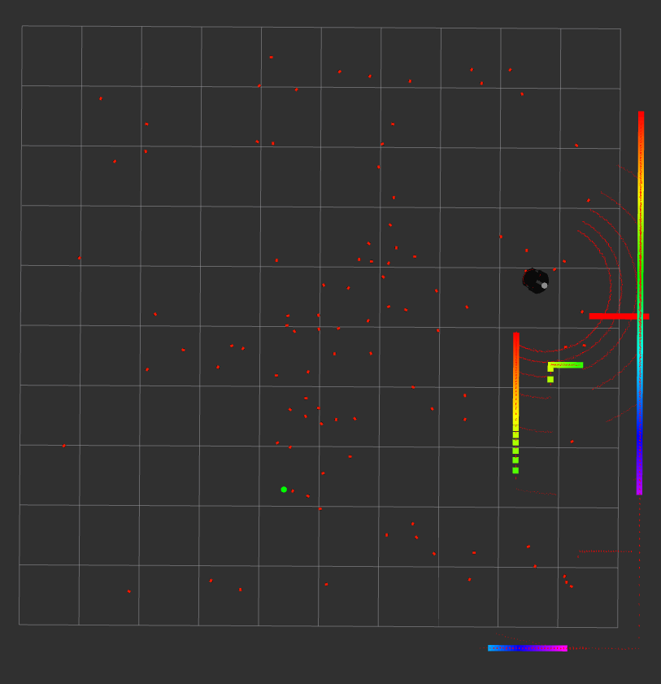}
    \caption{The RViz interface of the DQN and TD3 training process.}
    \label{fig:placeholder}
\end{figure}

High-level DQN generates subgoal markers in RViz, but the agent often rotates in place at the beginning of episodes. Training often terminates early or fails to converge, preventing meaningful quantitative comparison with TD3.

\subsection{Preliminary Analysis of DQN+TD3 Instability}

\begin{itemize}
    \item Multi-level non-stationarity: High-level and low-level policies update simultaneously, altering each other’s targets.
    \item Reward misalignment: Unified reward may produce conflicting gradients or incorrect layer attribution if not properly decomposed.
    \item Hyperparameter/scheduling mismatch: DQN and TD3 learning rates, update frequency, or replay buffer settings may be inconsistent.
    \item Environment/configuration sensitivity: High-level action discretization and environment parameters may hinder subgoal learning.
    \item Reward function parameter tuning: Current reward weights may not sufficiently balance high-level and low-level objectives.
\end{itemize}

\subsection{Limitations}

\begin{itemize}
    \item Validation scope: Only TD3 baseline has stable and repeatable quantitative results. DQN+TD3 hierarchical configuration is currently unstable.
    \item Training overhead: TD3 requires thousands of episodes to achieve stable behavior.
    \item Limited hyperparameter search: Only a small set of hyperparameters and schedules have been tried; broader automated search has not been conducted.
\end{itemize}

\subsection{Future Work}

\textbf{Our Next Steps:} Stabilize DQN+TD3 through systematic tuning of reward function, hyperparameters, and high-low layer interaction. Once stable, perform quantitative comparison with TD3 (success rate, collision rate, path efficiency, time cost) and statistical significance tests.

\textbf{Future Research Directions:} Building upon its hierarchical structure and inherent scalability, the framework is well-suited for extension to multi-robot coordination and complex navigation tasks. It holds strong potential for applications in logistics, surveillance, and search-and-rescue in high-dimensional and 3D environments.

\section{Conclusion}

This study evaluated the performance of the TD3 algorithm and the hierarchical DQN+TD3 framework in a ROS+Gazebo navigation simulation. TD3 alone demonstrated stable learning, effective convergence, and reliable navigation behavior. Although the DQN+TD3 hierarchical framework showed qualitative potential, it currently remains unstable and requires further tuning before meaningful quantitative evaluation. Future work will focus on stabilizing the DQN+TD3 framework and extending hierarchical reinforcement learning to multi-agent coordination and real-world deployment scenarios.

\bibliographystyle{unsrt}          
\bibliography{references}

\end{document}